\documentclass[iicol,sn-basic,pdflatex]{sn-jnl}

\usepackage{graphicx}%
\usepackage{multirow}%
\usepackage{amsmath,amssymb,amsfonts}%
\usepackage{amsthm}%
\usepackage{mathrsfs}%
\usepackage[title]{appendix}%
\usepackage{xcolor}%
\usepackage{textcomp}%
\usepackage{manyfoot}%
\usepackage{booktabs}%
\usepackage{algorithm}%
\usepackage{algorithmicx}%
\usepackage{algpseudocode}%
\usepackage{listings}%

\newcommand{\xiugai}[1]{\textcolor{black}{#1}}

\hypersetup{
    colorlinks=false,
    pdfborder={0 0 0},
}

\definecolor{blue}{rgb}{0,0,0}

\theoremstyle{thmstyleone}%
\theoremstyle{thmstyletwo}%

\theoremstyle{thmstylethree}%

\begin{document}

\title[Article Title]{LLM-powered Query Expansion for Enhancing Boundary Prediction in Language-driven Action Localization}


\author[1]{\fnm{Zirui} \sur{Shang}}\email{shangzirui@bit.edu.cn}

\author[1,2]{\fnm{Xinxiao} \sur{Wu}}\email{wuxinxiao@bit.edu.cn}

\author*[2]{\fnm{Shuo} \sur{Yang}}\email{yangshuo@smbu.edu.cn}

\affil[1]{\orgdiv{Beijing Key Laboratory of Intelligent Information Technology, School of Computer Science \& Technology}, \orgname{Beijing Institute of Technology}, \orgaddress{\country{China}}}

\affil[2]{\orgdiv{Guangdong Laboratory of Machine Perception and Intelligent Computing}, \orgname{Shenzhen MSU-BIT University}, \orgaddress{\country{China}}}



\abstract{
    Language-driven action localization in videos requires not only semantic alignment between language query and video segment, but also prediction of action boundaries. 
    However, the language query primarily describes the main content of an action and usually lacks specific details of action start and end boundaries, which increases the subjectivity of manual boundary annotation and leads to boundary uncertainty in training data. 
    In this paper, on one hand, we propose to expand the original query by generating textual descriptions of the action start and end boundaries through LLMs, which can provide more detailed boundary cues for localization and thus reduce the impact of boundary uncertainty. 
    On the other hand, to enhance the tolerance to boundary uncertainty during training, we propose to model probability scores of action boundaries by calculating the semantic similarities between frames and the expanded query as well as the temporal distances between frames and the annotated boundary frames. They can provide more consistent boundary supervision, thus improving the stability of training. 
    Our method is model-agnostic and can be seamlessly and easily integrated into any existing models of language-driven action localization in an off-the-shelf manner. Experimental results on several datasets demonstrate the effectiveness of our method.
}




\maketitle

\section{Introduction}\label{sec1}

Language-driven action localization, also known as video moment retrieval~\citep{gao2021fast,wu2021diving,zeng2021multi,tang2021frame} or temporal sentence grounding~\citep{yuan2019semantic,liu2021context,liu2021progressively}, aims to localize the temporal action segment semantically aligned with a natural language query in an untrimmed video. 
It has received increasing attention in recent years and has wide applications in many downstream tasks such as video question answering~\citep{sun2021video} and video editing~\citep{zhang2022ai}. 
Recent studies~\citep{gao2017tall,liu2023jointly,liu2021adaptive,zhang2020learning} have achieved remarkable progress in language-driven action localization by aligning the video segment with the language query~\citep{ghosh2019excl,zhang2023text,liu2022memory,liu2023towards} and predicting the desired action boundaries using rigid boundary annotations as supervision~\citep{zhang2020span,liu2022skimming,liu2022reducing,liu2021context}.

\begin{figure}[t]
\centering  
    \includegraphics[width=\linewidth]{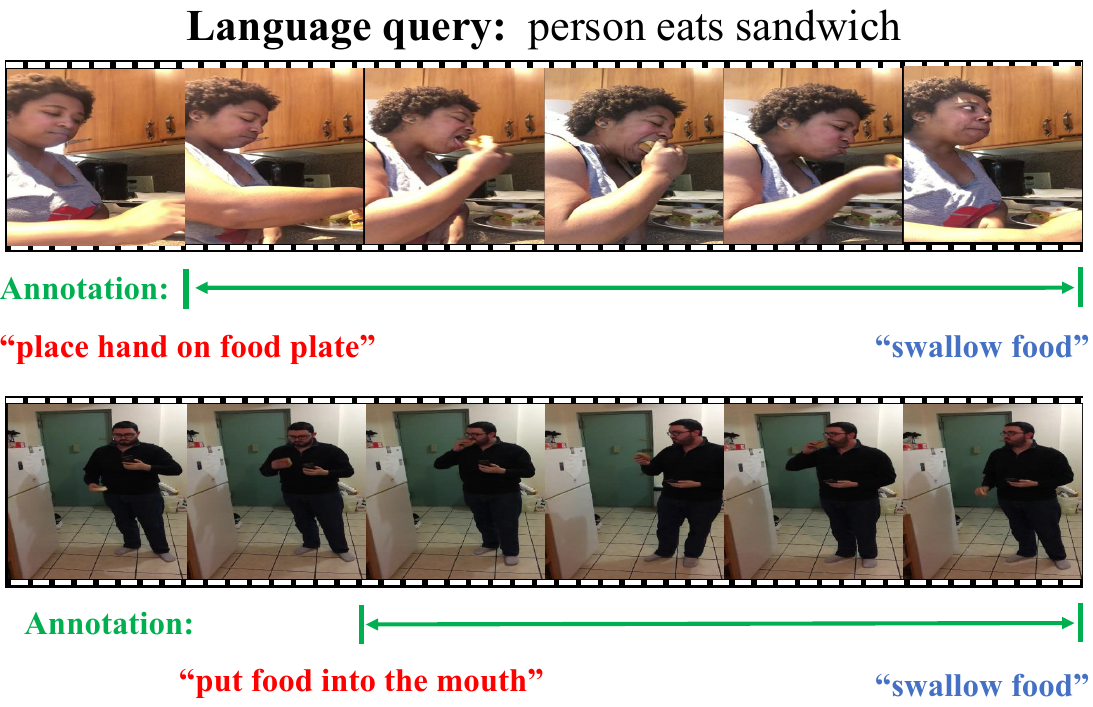}
    \caption{Illustration of boundary uncertainty for the same language query. The orange text represents the motion of the start boundary annotation and the blue text represents the motion of the end boundary annotation.}
    \label{fig:intro}
\end{figure}




However, the language query primarily describes the main content of an action, and lacks details of start and end boundaries. Different human annotators may have different subjective interpretations of action boundaries, and give inconsistent boundary annotations for the same type of action in different videos, resulting in boundary uncertainty. 
As shown in Figure~\ref{fig:intro}, for the language query of ``person eats sandwich'' that describes the same action ``eat'', the start boundary annotation of one video (a) refers to the motion ``place hand on food plate", which that of another video (b) refers to the motion ``put food into the mouth". 
Therefore, learning to localize accurate boundary timestamps from inconsistent boundary annotations is highly challenging and is still under exploration.

To address this challenge, we propose LLM-powered query expansion, which leverages a large language model (LLM) to generate detailed textual descriptions of action start and end boundaries to expand the original query. Specifically, we first design several prompts, such as \textit{“Please describe the beginning and ending process in one sentence of the following action [ACTION],
The description you generate cannot contain any objects that are not present in the action”}, for the LLM to generate fine-grained descriptions about motions, poses, or temporal differences that distinguish boundaries.
\xiugai{Then, we propose a query-guided temporal modeling module that incorporates the expanded query into visual representation learning. By injecting explicit boundary cues, the model is able to extract discriminative features that capture boundary details the original query might overlook, thereby refining the ``input'' perception of action boundaries and reducing the impact of uncertainty.}

\xiugai{On the other hand, to further enhance the tolerance to boundary uncertainty during training, we propose a boundary probability modeling module that estimates the probability of each video frame being an action boundary based on the expanded query. This provide the ``output'' supervision with probabilistic flexibility, enabling the model to map discriminative features to action boundaries.}
Specifically, for each frame, we model the probability of being an action boundary by calculating its semantic similarity to the expanded query and its temporal distance to the labeled boundary frame. The semantic similarity ensures that frames relevant to the action described by the expanded query are prioritized, while the temporal distance reflects the intuition that frames closer to the annotated boundary are more likely to correspond to the action boundary.
The probability scores, instead of rigid annotations, reduce over-reliance on a single annotated boundary by providing more flexible and reliable supervision, thereby better capturing motion-related information near the boundary.
In addition, the clear guidance of the expanded query ensures that the probability scores are more consistent across similar actions, further providing more consistent supervision and improving the stability of training.


The main contributions can be summarized as follows: 
 \begin{itemize}
     \item We propose a query expansion method that expands the language query through LLM to capture more details of action boundaries, effectively reducing the impact of boundary uncertainty in action localization.
     \item We propose a boundary probability modeling module that utilizes the expanded query to transform rigid boundary annotations into probability scores for training, successfully improving the robustness to boundary uncertainty.
     \item Our method can be seamlessly and easily integrated into existing models in an off-the-shelf manner. Extensive experiments on five state-of-the-art models across three benchmark datasets demonstrate the effectiveness of our method.
 \end{itemize}

\section{Related Work}\label{sec2}

Existing methods of language-driven action localization can be broadly categorized into two groups: proposal-based methods and proposal-free methods.
The proposal-based methods first generate segment proposals using sliding windows, proposal generation, or anchor-based methods and then rank them according to the query. Through exhausting proposal generation and directly aligning the semantics of the query and proposals, the proposal-based methods achieve better performance. 
Early methods such as CTRL~\citep{gao2017tall} and MCN~\citep{hendricks2017localizing} use sliding windows of varying scales to generate candidate video segments, while later methods such as SPN~\citep{xu2017r}, QSPN~\citep{xu2019multilevel}, and SAP~\citep{chen2019semantic} adopt segmented proposal networks tailored to specific queries. Recent studies, including 2D-TAN~\citep{zhang2020learning}, RaNet~\citep{gao2021relation}, and CDN~\citep{wang2022cross}, utilize two-dimensional feature maps to model relationships across segments of different durations.
MIPGN~\citep{fang2025multi} proposes a multi-modal integrated proposal generation network to guide adaptive proposal generation, incorporates object and motion tags for richer multi-modal representations, and employs dynamic negative proposal mining to improve retrieval accuracy.
VGCI~\citep{lv2025variational} introduces a variational inference framework for weakly supervised video moment retrieval, modeling a global clue as a latent Gaussian variable to guide query reconstruction and enhance video-query alignment.

The proposal-free methods first learn cross-modal interactions between query and video, and then predict the frame-wise boundary probabilities ~\citep{zhang2020span,liu2022skimming,liu2022reducing,liu2021context} or regress the start and end boundary frames of the target segment~\citep{ghosh2019excl,zhang2023text,liu2022memory,liu2023towards}. Zhang~\textit{et al.}~\citep{zhang2021natural} search for the target action within a highlighted region in a span-based question-answering framework. 
Mun~\textit{et al.}~\citep{mun2020local} extract the implicit semantic information from local to global by a sequential query attention module. Sun~\textit{et al.}~\citep{Sun2022YouNT} leverage fine-grained intra-modality clues to explore deeper inter-modality information. 
UniVTG-NA~\citep{flanagan2025moment} introduces a negative-aware video moment retrieval framework that distinguishes between in-domain and out-of-domain negative queries, enabling robust rejection of irrelevant queries.
SSTr~\citep{huo2025skim} proposes a Skim-and-Scan Transformer architecture, leveraging dual-granularity feature interaction and dual-supervision strategies to model cross-video semantics at multiple temporal scales.
Recently, DETR-based methods~\citep{xiao2024bridging,yang2024task,jang2023knowing,moon2023correlation} have been proposed to apply the popular DETR~\citep{carion2020end} architecture into language-driven action localization, which have achieved remarkable performance. 
BM-DETR~\citep{jung2025background} proposes a background-aware moment detection transformer that leverages contrastive learning with negative queries to better utilize surrounding context and alleviate weak alignment issues in video moment retrieval.
W2W~\citep{liu2025and} proposes a progressive ``what and where" framework, combining cross-modal semantic alignment via Initial Semantic Projection and Clip Semantic Mining with moment-level temporal context modeling using a dual-branch Multi-Context Perception module.

The most related work to our method are ~\citep{otani2020uncovering}, ~\citep{pan2022video} and \cite{huang2022video}, which focuses on solving the problem of boundary annotation uncertainty. 
Otani~\textit{et al.}~\citep{otani2020uncovering} quantitatively study the uncertainty in the temporal annotations by collecting multiple boundaries for the same action from different annotators, but do not explicitly propose a solution. 
Pan~\textit{et al.}~\citep{pan2022video} adopt Gaussian smoothing to generate soft labels for each boundary, and Huang~\textit{et al.}~\citep{huang2022video} propose elastic moment bounding between annotated boundaries and predicted boundaries to generate pseudo labels. 
These methods heuristically expand the rigid start and end boundaries into soft labels, considering only temporal visual differences. 
In this paper, we leverage a powerful LLM to exploit more explicit interpretations for each boundary, thereby alleviating the ambiguity of boundary definitions and reducing the uncertainty of boundary annotations.

\begin{figure*}[t]
    \centering
    \xiugai{
    \includegraphics[width=\linewidth]{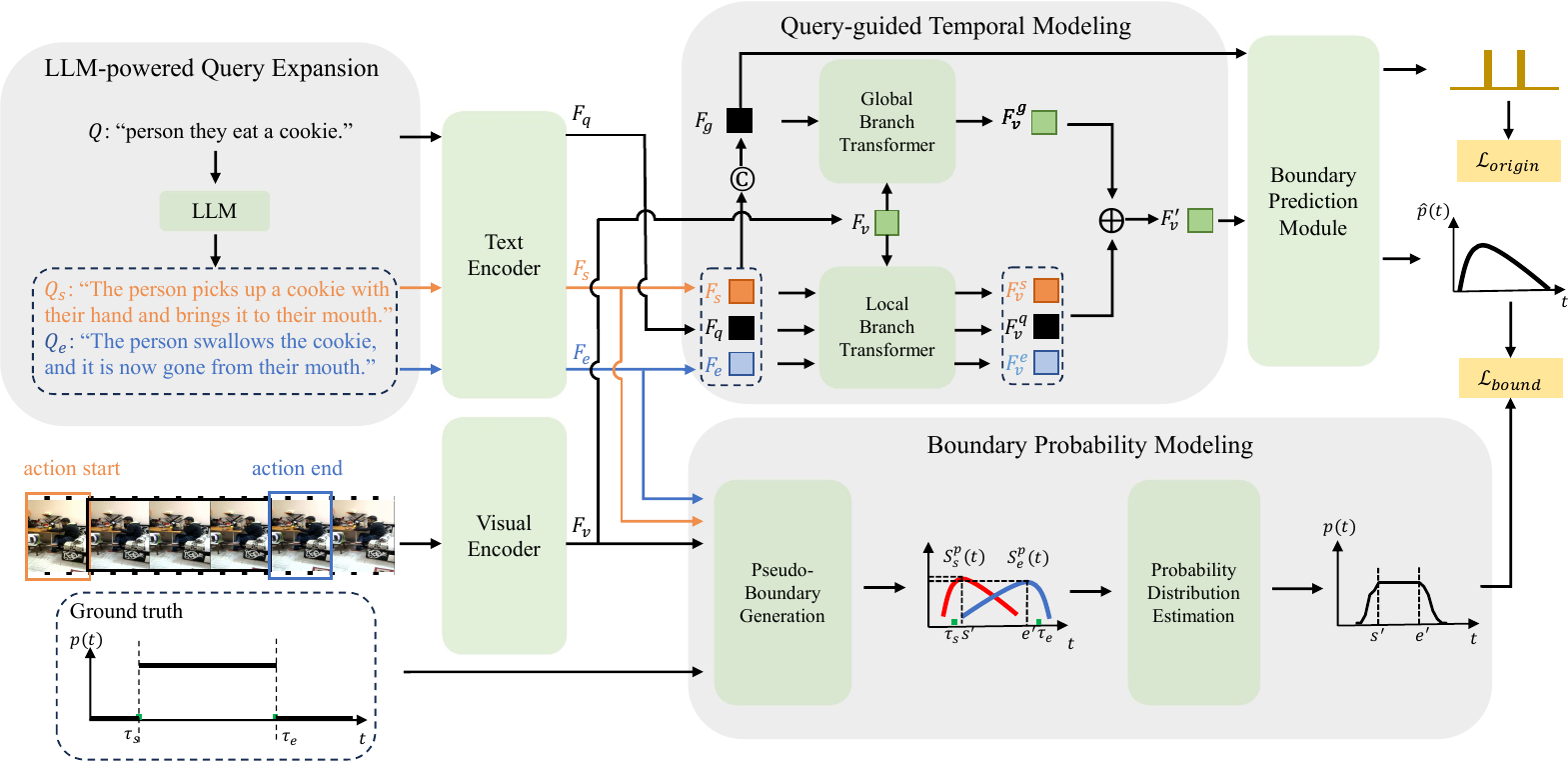}
    \caption{Overview of the proposed action localization framework. The pipeline first leverages a large language model to perform text expansion, generating detailed sentences that describe the beginning and ending stages of the target action based on the original content-focused query. These expanded boundary cues, along with the visual video, are processed by the base model to obtain initial text and video features. To refine the multimodal representations, we introduce a query-guided temporal modeling module that fuses the detailed start and end descriptions with the video features, employing global and local branches to capture holistic context and stage-specific boundary details. Finally, a boundary probability modeling module is proposed to estimate the likelihood of action boundaries across the temporal sequence, converting rigid ground-truth intervals into soft probability scores for more robust supervision during training.}
    \label{fig:pipeline}
    }
\end{figure*}

\section{Our Method}\label{sec3}

\xiugai{In this section, we present the proposed framework for language-driven action localization under boundary uncertainty. We first provide a high-level Overview of the pipeline in Sec. 3.1. Sec. 3.2 defines the Base Model we use. To enrich the semantic boundary cues, the LLM-powered Query Expansion process is introduced in Sec. 3.3. The proposed Query-guided Temporal Modeling and Boundary Probability Modeling are introduced in Sec. 3.4 and 3.5, respectively.}

\subsection{Overview}
Given an untrimmed video $V$ and a language query $Q$, we present the video as $V=\{v_t\}^{T}_{t=1}$, where $v_t$ is the $t$-th frame, and $T$ is the number of total frames. The language query with $N$ words is denoted as $Q=\{q_n\}^{N}_{n=1}$. Our task aims to locate the start and end timestamps $(\tau_s, \tau_e)$ of the action described by the language query $Q$ in the untrimmed video $V$. 

We first prompt the LLM to generate fine-grained textual descriptions of start and end boundaries as expanded queries, denoted as $Q_s=\{q_n\}^{N_s}_{n=1}$ and $Q_e=\{q_n\}^{N_e}_{n=1}$, where $N_s$ and $N_e$ are the lengths of the start and end queries, respectively.
We then extract the video \xiugai{features} $F_v \in \mathbb{R}^{T \times D}$ for $V$, and extract the start query \xiugai{features} $F_s \in \mathbb{R}^{N_s \times D}$, the original query \xiugai{features} $F_q \in \mathbb{R}^{N \times D}$, the end query \xiugai{features} $F_e \in \mathbb{R}^{N_e \times D}$ for $Q_s$, $Q$, and $Q_e$, respectively, using the visual and text encoders of the base model \xiugai{$\mathcal{M}$}.
\xiugai{We propose a query-guided temporal modeling module to get enhanced query features $F_g \in \mathbb{R}^{N \times D}$ and visual features $F'_v \in \mathbb{R}^{N \times D}$}, which are then fed into the localization head of \xiugai{$\mathcal{M}$} to obtain predictions. We also design a boundary probability modeling module to generate probability scores of action boundaries by combining cross-modal semantic similarity and inter-frame temporal distance to refine the predictions.

\xiugai{
\subsection{Base Model}
\label{sec:base_model}
Our proposed method is designed to be model-agnostic. To demonstrate its effectiveness, we build upon a kind of representative DETR-based framework\citep[e.g.,][]{lei2021detecting,moon2023query}, as our base model. For structural clarity, this framework is abstracted into two core components consistent with our overall pipeline: Visual and Text Encoders for input representation, and a \xiugai{Localization Head} for task-specific outputs. In our full pipeline, the Query-guided Temporal Modeling module (Sec 3.4) is integrated as an intermediate enhancement between these two components to refine the feature representations.
}

\noindent \xiugai{
\textbf{Visual and Text Encoder}.
Following the problem formulation, we first obtain raw video and query features using pretrained backbones. Specifically, $F_v^{raw}$ is formed by concatenating SlowFast~\citep{feichtenhofer2019slowfast} and CLIP~\citep{radford2021learning} visual features, while $\{F_q, F_s, F_e\}^{raw}$ are extracted via the CLIP text encoder. To bridge the dimensionality gap between these heterogeneous backbones and the subsequent Transformer, we then employ 2-layer MLPs as projection layers to map all raw features into a unified hidden dimension $D$:
\begin{equation}
F_i = \text{MLP}_i(F_i^{raw}), \quad i \in \{v, q, s, e\}.
\end{equation}
}


\noindent \xiugai{
\textbf{Localization Head}. This module adopts a Transformer-based encoder-decoder architecture to process the interacted features and predict temporal boundaries. 
In the base model, the projected video features $F_v$ and query features $F_q$ are integrated by simply concatenating them along the sequence length dimension to form a joint input sequence:
\begin{equation}
E_{input} = [F_v; F_q] \in \mathbb{R}^{(T+N) \times D}.
\end{equation}
The input features $E_{input}$ are then fed into a stack of $T$ Transformer encoder layers, which utilize multi-head self-attention to generate contextualized representations $E_{enc} \in \mathbb{R}^{(T+N) \times D}$. The decoder then refines $M$ learnable moment queries through cross-attention over $E_{enc}$, producing moment-aware embeddings $E_{dec} \in \mathbb{R}^{M \times D}$.
}

\xiugai{
To generate final predictions, $E_{enc}$ is used to compute clip-level saliency scores $S \in \mathbb{R}^{T}$ via a linear layer. Simultaneously, $E_{dec}$ is passed through task-specific heads: a 3-layer FFN to regress normalized temporal boundaries $\hat{m} = (\hat{c}, \hat{w})$ and a linear classifier to produce foreground/background scores.
}

\noindent \xiugai{
\textbf{Training Loss}. To ensure a fair comparison and maintain model-agnosticism, our framework inherits the original optimization objectives of the chosen base model $\mathcal{M}$, as described in their respective works~\citep{lei2021detecting,moon2023query}. Specifically, the joint origin loss $\mathcal{L}_{origin}$ is formulated as a combination of three functional components:
\begin{equation}
\mathcal{L}_\text{origin} = \mathcal{L}_\text{loc} + \mathcal{L}_\text{cls} + \mathcal{L}_\text{sal},
\end{equation}
where $\mathcal{L}_{loc}$ regresses the temporal boundaries, $\mathcal{L}_{cls}$ performs foreground-background classification, and $\mathcal{L}_{sal}$ facilitates saliency ranking.
}

\subsection{LLM-powered Query Expansion}

\begin{figure*}
    \centering
    \includegraphics[width=\linewidth]{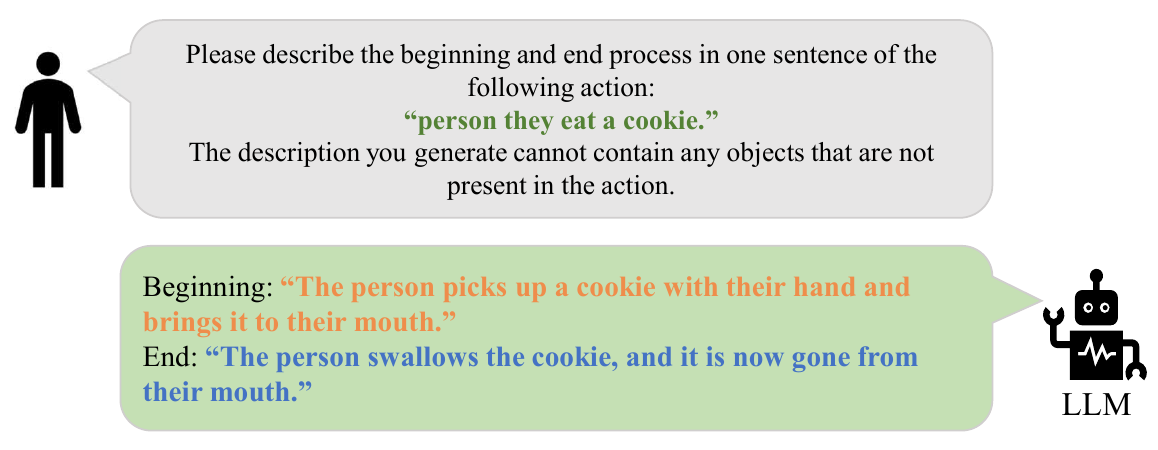}
    \caption{An example of the designed question and the corresponding answers from LLM.}
    \label{fig:questions}
\end{figure*}




In language-driven action localization, the input language query primarily describes the main content of an action and usually neglects details of start and end boundaries, which increases subjective interpretations by human annotators, leading to inconsistent boundary annotations. To reduce the impact of boundary uncertainty, we leverage LLMs to expand the original query by explicitly describing more detailed information of action boundaries, such as motions, human poses, and temporal differences that distinguish boundaries, providing more fine-grained boundary cues for localization.

We design a prompt for the LLM (LLaMa3-8B~\citep{dubey2024llama} is used in this paper) to generate two expanded queries: one query describing the action start and the other describing the action end. The prompt is crafted to guide the LLM in producing descriptions that are semantically aligned with the original query and explicitly focus on the temporal boundaries of action, such as \textit{“Please describe the beginning and ending process in one sentence of the following action [ACTION]”}. To reduce the potential hallucination problem of LLMs and maintain the consistency of the expanded queries for similar actions, we instruct the LLM: \textit{“The description you generate cannot contain any objects that are not presented in the action”}. 
\xiugai{Furthermore, to ensure conciseness and minimize textual noise, we require the LLM to describe each process in exactly one short sentence and strictly limit each sentence to no more than 20 words. 
This ensures that the expanded queries align with the semantic content described by the original query while providing detailed boundary-specific information.}  Figure~\ref{fig:questions} provides an example of the designed \xiugai{prompt} and the corresponding outputs.

By providing finer boundary details and enriching the semantic context for localization, the expanded queries help capture fine-grained temporal information that is overlooked by the original coarse query.
For the expanded start query $Q_s$ and end query $Q_e$, we use the text encoder of the base model to extract their corresponding feature $F_s$ and $F_e$, respectively, where $D$ is the feature dimension.

\subsection{Query-guided Temporal Modeling}
\label{sec:qtm}

\xiugai{To effectively leverage the semantic guidance from expanded queries, we propose the Query-guided Temporal Modeling module. As illustrated in the overall pipeline, the module acts as a plug-and-play enhancement layer inserted before the Localization Head. By injecting explicit boundary cues into the visual representation learning, this module is able to extract discriminative features that capture boundary details the original query might overlook, thereby refining the ``input'' perception of action boundaries. Specifically, the module employs a dual-branch architecture: local branch that focuses on fine-grained boundary representations at specific action stages, and a global branch that captures the overall temporal structure of the entire sequence.}



\noindent \xiugai{
\textbf{Local Branch.} 
The local branch enhances the video feature $F_v \in \mathbb{R}^{T \times D}$ by performing fine-grained temporal modeling for each action stage, using the start query feature $F_s \in \mathbb{R}^{N_s \times D}$, the original query feature $F_q \in \mathbb{R}^{N_q \times D}$, and the end query feature $F_e \in \mathbb{R}^{N_e \times D}$ as guidance, respectively. Specifically, $F_v$ is treated as the \textit{query}, while $F_s$, $F_q$, or $F_e$ serves as the \textit{key} and \textit{value} in the Multi-Head Cross-Attention (MHCA) mechanism, resulting in locally enhanced features $F^s_v, F^q_v, F^e_v \in \mathbb{R}^{T \times D}$:
\begin{equation}
F_v^i = \text{MHCA}(F_v, F_i, F_i), \quad i \in {s, q, e},
\end{equation}
which encode temporal details specific to the start, main body, and end stages of the action.
}

\noindent \xiugai{
\textbf{Global Branch.}
The Global Branch enhances $F_v$ by capturing holistic temporal dependencies across the entire sequence, using the concatenated query feature $F_{g}=[F_s;F_q;F_e] \in \mathbb{R}^{(N_s+N_q+N_e) \times D}$ as guidance. Specifically, $F_v$ serves as the \textit{query}, and $F_{g}$ is used as \textit{key} and \textit{value} to obtain the globally enhanced feature $F^g_v \in \mathbb{R}^{T \times D}$:
\begin{equation}
F_v^g = \text{MHCA}(F_v, F_g, F_g).
\end{equation}
}

Finally, the globally enhanced video feature and the locally enhanced video features are weighted fused to obtain the final video feature $F'_v$:
\begin{equation} \label{equ:enhance}
    F'_v = a F^g_v + b (F^s_v+F^q_v+F^e_v),
\end{equation}
where $a$ and $b$ are hyper-parameters. 

\xiugai{
Through this dual-branch refinement, the initial features $F_v$ and $F_q$ are transformed into boundary-discriminative visual features $F'_v$ and enhanced query features $F_g$. While the base model directly utilizes the raw concatenated features $E_{input} = [F_v; F_q]$ (Eq. 2), the integrated framework replaces it with the enhanced joint sequence $E'_{input} = [F'_v; F_g]$ as the final input for the subsequent Localization Head.
}

\subsection{Boundary Probability Modeling}
Although the expanded queries effectively reduce boundary uncertainty by providing detailed boundary descriptions, they cannot completely eliminate the uncertainty caused by human annotators' subjective interpretations of action boundaries.
\xiugai{
To this end, we propose a boundary probability modeling module that leverages the rich semantic information from expanded queries to estimate probability scores of potential action boundaries. By using these scores as flexible ``output'' supervision instead of rigid annotations, the model is able to focus on a boundary region rather than a single frame to map the discriminative features to action boundaries. This provides the probabilistic flexibility to effectively enhance the tolerance to boundary uncertainty during training.}
In addition, the probability scores of boundaries are more consistent between similar actions, further providing more consistent supervision and improving the stability of training.



The boundary probability modeling module consists of two steps: a pseudo boundary generation step that identifies the most confident action boundaries relevant to the expanded queries, and a probability score estimation step that assigns likelihoods to frames based on their relevances to the pseudo boundaries and the expanded queries, converting rigid annotations into probabilities as supervision for training.

\noindent\textbf{Pseudo Boundary Generation.} 
Since the action description in the expanded query may occur before or after the annotated boundaries, we introduce pseudo boundaries around the ground-truth boundaries, based on two key factors: temporal distance and semantic similarity, to identify the most confident boundaries. The temporal distance emphasizes frames close to the ground-truth boundaries, and the semantic similarity prioritizes frames that align with the boundaries described by the expanded query.

Specifically, for each frame $i$, we calculate its scores $S_s^p(i)$ and $S_e^p(i)$ by 
\xiugai{
\begin{equation} \label{equ:pseudo} 
    \begin{aligned}
       S_s^p(i) = \text{sim}(F_{v,i},F_s)- \alpha \cdot\text{dis}(i,\tau_s), \\
       S_e^p(i) = \text{sim}(F_{v,i},F_e)- \alpha \cdot\text{dis}(i,\tau_e),
    \end{aligned}
\end{equation}
}
where $F_{v,i}$ represents the visual feature of the $i$-th frame, $F_s$ and $F_e$ represent the start query feature and the end query feature, respectively, $\text{sim}(,\cdot,)$ represents the semantic similarity calculated by cosine similarity, $\text{dis}(,\cdot,)$ represents the temporal distance, normalized by the total video length $T$, \textit{i.e.}, $\text{dis}(i,\tau_s)=|i-\tau_s|/T$
\xiugai{, and $\alpha$ is a hyper-parameter to balance two terms}.
Then, the pseudo boundaries are determined by maximizing the scores:
\xiugai{
\begin{equation}
\begin{aligned}
\tau_s' = \operatorname{arg\,max}_{i}S_s^p(i), \\
\tau_e' = \operatorname{arg\,max}_{i}S_e^p(i).
 \end{aligned}
\end{equation}
}

\noindent\textbf{Probability Score Estimation.} 
Using the pseudo boundaries \xiugai{$(\tau_s', \tau_e')$}, we convert rigid boundary annotations into probability scores to better capture boundary uncertainty. Frames that are closer to the pseudo boundaries and show higher semantic similarities to the expanded queries are assigned with higher probabilities of being action boundaries.

Specifically, for each frame \xiugai{$i < \tau_s'$}, we calculate its score $ S_s(i)$ based on its temporal distance to the pseudo boundary \xiugai{$\tau_s'$} and its semantic similarity to the start query $Q_s$:
\xiugai{
\begin{equation} \label{equ:score_start}
    S_s(i) = \text{sim}(F_{v,i},F_s) - \alpha \cdot \text{dis}(i,\tau_s').
\end{equation}
}
Based on these scores, we generate a probability score that focuses on the frames that are most likely to be action boundaries. To achieve this, we first filter out frames with scores below a predefined threshold $\tau$, keeping only those frames with high relevance to the action boundary:
\begin{equation} \label{equ:filter}
    S^{'}_{s}(i)=\begin{cases}
            0, & S_{s}(i) < \tau, \\
            S_{s}(i), & S_{s}(i) \geq \tau. \\
    \end{cases}
\end{equation}
Then, we generate a probability score of the $i$-th frame being the start boundary by applying min-max normalization to $ S^{'}_{s}(i)$:
\begin{equation} \label{equ:start_prob}
    p_s(i) = \begin{cases} 
      \dfrac{S_s'(i) - S_{s,\min}'}{S_{s,\max}' - S_{s,\min}'}, & S_s'(i) > 0, \\
      0, & S_s'(i) = 0,
   \end{cases}
\end{equation} 
where $S_{s,\max}' = \max \{ S_s'(i) \mid S_s'(i) > 0 \}$ and $S_{s,\min}' = \min \{ S_s'(i) \mid S_s'(i) > 0 \}$. The same process is applied to generate a probability score $p_e(i)$ of the $i$-th frame being the end boundary, where \xiugai{$i>\tau_e'$}.

Finally, the rigid boundary annotations are \xiugai{converted} into probability scores by
\xiugai{
\begin{equation} \label{equ:prob}
    \boldsymbol{p}(i) = \begin{cases} 
      p_s(i), & 0 \leq i < \tau_s', \\
      1, & \tau_s' \leq i \leq \tau_e', \\
      p_e(i), & \tau_e' < i \leq T. 
   \end{cases}
\end{equation}
}

\noindent \xiugai{\textbf{Loss Function.} We feed the enhanced video feature $F^{'}_{v}$ and the query feature $F_{g}$ into the localization head of the base model $\mathcal{M}$, denoted as $\mathcal{M}_{lh}$, to predict the probability $\boldsymbol{\hat{p}}(i)$ of a frame $i$ being in the action segment:
\begin{equation} \label{equ:predict}
       \boldsymbol{\hat{p}}(i) = \mathcal{M}_{lh}(F^{'}_{v}, F_{g}).
\end{equation}
During training, we introduce a boundary loss to minimize the discrepancy between the predicted probability score $\boldsymbol{\hat{p}}(i)$ and the generated probability score $\boldsymbol{p}(i)$ in Eq.~(\ref{equ:prob}), using the cross-entropy loss over all frames:
\begin{equation} \label{equ:bound_loss}
    \mathcal{L}_\text{bound} = \Sigma_{i=1}^{T} \text{Cross-Entropy}(\boldsymbol{p}(i),\boldsymbol{\hat{p}}(i)).
\end{equation}
The final loss function is computed as 
\begin{equation} \label{equ:loss}
    \mathcal{L} = \mathcal{L}_\text{bound} + \mathcal{L}_\text{origin},
\end{equation}
where $\mathcal{L_\text{origin}}$ is the original loss that the base model uses.}

\section{Experiment}\label{sec4}

\xiugai{
In this section, we conduct extensive experiments to evaluate the performance of our proposed method. We first introduce the Datasets and Experimental Settings in Sec. 4.1 and 4.2, respectively. A comprehensive Comparison with existing methods is presented in Sec. 4.3 to demonstrate our competitive results. To analyze the contribution of each component, we perform detailed Ablation Studies in Sec. 4.4. Furthermore, a User Study on the quality of generated queries is provided in Sec. 4.5 to assess the effectiveness of our LLM-powered expansion. We also provide Qualitative Results in Sec. 4.6 to visualize the localization performance of our framework. Finally, the efficiency analysis is presented in Sec. 4.7.
}

\subsection{Datasets}
\noindent \textbf{Qvhighlights}~\citep{lei2021detecting} is a recently introduced dataset that supports both language-driven action localization and highlight detection. The dataset contains 10,148 videos from YouTube, covering a wide range of content, and includes 10,310 text queries associated with 18,367 annotated moments, with an average of 1.8 moments per query. Different from other datasets that typically have one-to-one query-action pairs, QVHighlights provides a more complex real-world scenario with multiple actions per query. Since the official test set does not have ground-truth annotations, the testing is performed on the official validation set.

\noindent \textbf{Charades-STA}~\citep{gao2017tall} is constructed by extending the Charades dataset~\citep{sigurdsson2016hollywood} for language-driven action localization. It contains 6,670 videos and 16,124 queries, where 12,404 query-action pairs are used for training and 3720 for testing. The average video duration is 30.59 seconds, and each video contains 2.41 annotated moments, and the moment has an average duration of 8.09 seconds.

\noindent \textbf{TACoS}~\citep{regneri2013grounding} consists of videos selected from the MPII Cooking Composite Activities video corpus~\citep{rohrbach2012script}. It contains 18,818 video-query pairs of different cooking activities. Each video contains 148 queries on average, some of which are annotations of short video clips.

\subsection{Experimental Settings}
\noindent \textbf{Evaluation Metrics.}
Following the previous work~\citep{moon2023query,taichi2024emnlp}, we adopt Recall1@IoU $\mu$ and mean average precision (mAP) as evaluation metrics: 
(1) R1@$\mu$, which denotes the percentage of top-1 predicted moment having IoU with ground truth larger than threshold $\mu$. 
(2) mAP, which denotes the mean Average Precision and evaluates the overall performance by averaging the precision values of different IoU thresholds. 

\xiugai{
\noindent \textbf{Implementation Details.}
We use LLaMa3-8B~\citep{dubey2024llama} as the LLM to generate expanded queries.
Both the Local and Global branches utilize a single-layer Transformer with 8 attention heads.
The hyper-parameter $\alpha$ is set to 1, while $a$ and $b$ in Eq.(\ref{equ:enhance}) are both set to 1. The threshold $\tau$ in Eq.(\ref{equ:filter}) is set to 0.8.
To ensure a fair comparison, other training configurations (e.g., learning rate, optimizer, and batch size) strictly follow the default settings of the respective base models.
We conduct experiments on a single NVIDIA RTX 4090 GPU.
}

\begin{table*}[t]
\setlength\tabcolsep{2pt}
\centering
\caption{Results of different existing methods integrated with our modules on the Qvhighlights, Charades-STA and TACoS datasets.}
\label{tab:fully}
\begin{tabular}{l|ccc|ccc|ccc}
     \hline
     \hline
     \multicolumn{1}{l|}{\multirow{2}{*}{Method}} & \multicolumn{3}{c|}{Qvhighlights} & \multicolumn{3}{c|}{Charades-STA} & \multicolumn{3}{c}{TACoS} \\
     & R1@0.5 & R1@0.7 & mAP & R1@0.5 & R1@0.7 & mAP & R1@0.5 & R1@0.7 & mAP\\
     \hline
     QD-DETR~\citep{moon2023query} & 63.16 & 46.39 & 40.39 & 58.17 & 36.94 & 34.75 & 38.17 & 20.72 & 20.20\\
     QD-DETR+Ours & \textbf{64.13} & \textbf{47.10} & \textbf{41.16} & \textbf{59.71} & \textbf{38.58} & \textbf{35.97} & \textbf{40.83} & \textbf{24.04} & \textbf{22.49}\\ 
     \hline
     Eatr~\citep{jang2023knowing} & 58.26 & 41.61 & 36.58 & 55.67 & 33.28 & 33.95 & 31.69 & 16.27 & 16.92\\
     Eatr+Ours & \textbf{60.13} & \textbf{43.23} & \textbf{37.91} & \textbf{56.83} & \textbf{33.95} & \textbf{34.64} & \textbf{34.12} & \textbf{16.92} & \textbf{18.04}\\
     \hline
     TaskWeave~\citep{yang2024task} & 64.26 & 48.17 & 44.20 & 56.85 & 34.44 & 35.37 & 38.52 & 20.82 & 19.99 \\
     TaskWeave+Ours & \textbf{65.31} & \textbf{50.00} & \textbf{44.90} & \textbf{58.44} & \textbf{36.51} & \textbf{36.84} & \textbf{39.20} & \textbf{21.5} & \textbf{20.70} \\
     \hline
     UVCOM~\citep{xiao2024bridging} & 64.45 & 49.81 & 43.31 & 57.18 & 35.43 & 35.60 & 40.20 & 23.30 & 22.11\\
     UVCOM+Ours & \textbf{65.31} & \textbf{51.23} & \textbf{44.27} & \textbf{58.84} & \textbf{36.83} & \textbf{36.60} & \textbf{40.71} & \textbf{24.59} & \textbf{23.05}\\
     \hline
     CG-DETR~\citep{moon2023correlation} & 65.48 & 50.97 & 44.03 & 57.71 & 35.61 & 35.07 & 39.79 & 24.89 & 23.28 \\
     CG-DETR+Ours & \textbf{66.52} & \textbf{52.19} & \textbf{44.60} & \textbf{58.87} & \textbf{36.77} & \textbf{36.33} & \textbf{40.27} & \textbf{25.31} & \textbf{23.77}\\
     \hline
     \hline
\end{tabular}
\end{table*}

\subsection{Comparison with existing methods}
We conduct experiments by integrating our modules into several existing methods (base models), including QD-DETR~\citep{moon2023query}, Eatr~\citep{jang2023knowing}, TaskWeave~\citep{yang2024task}, UVCOM~\citep{xiao2024bridging}, and CG-DETR~\citep{moon2023correlation}.
The comparison results on the Qvhighlights, Charades-STA, and TACoS datasets are shown in Table~\ref{tab:fully}. 
The results of the original methods are based on our reproduction following~\cite{taichi2024emnlp}.
From the results, we observe that all the methods integrated with our moduels consistently achieve better performance on all three datasets, highlighting the effectiveness of query expansion and probability score modeling in action boundary prediction. 

\subsection{Ablation Study}
To further evaluate different components of our method, we conduct ablation studies using QD-DETR as base model.


\begin{table*}[t]
\setlength\tabcolsep{2pt}
\centering
\xiugai{
\caption{Results of different proposed modules with QD-DETR across three datasets. ``QTM'' and ``BPM'' denote Query-guided Temporal Modeling and Boundary Probability Modeling, respectively.}
\label{tab:abl_full_component}
\begin{tabular}{cc|ccc|ccc|ccc}
\hline \hline
\multicolumn{2}{c|}{Module} & \multicolumn{3}{c|}{Charades-STA} & \multicolumn{3}{c|}{QVHighlights} & \multicolumn{3}{c}{TACoS} \\
QTM & BPM & R1@0.5 & R1@0.7 & mAP & R1@0.5 & R1@0.7 & mAP & R1@0.5 & R1@0.7 & mAP \\ \hline
 &  & 58.17 & 36.94 & 34.75 & 63.16 & 46.39 & 40.39 & 38.17 & 20.27 & 20.91 \\
\checkmark &  & 59.06 & 37.87 & 35.15 & 63.79 & 46.71 & 40.91 & 39.37 & 22.39 & 21.45 \\
 & \checkmark & 59.33 & 38.25 & 35.41 & 63.60 & 46.82 & 40.57 & 40.04 & 22.14 & 21.75 \\
\checkmark & \checkmark & \textbf{59.71} & \textbf{38.58} & \textbf{35.97} & \textbf{64.13} & \textbf{47.10} & \textbf{41.16} & \textbf{40.83} & \textbf{24.04} & \textbf{22.49} \\
\hline \hline
\end{tabular}
}
\end{table*}

\noindent \xiugai{\textbf{Evaluation of Different Modules.} 
To evaluate the effectiveness of the proposed modules, we conduct ablation studies across different datasets by removing them separately. The experimental results are shown in Table~\ref{tab:abl_full_component}, where QTM and BPM denote the query-guided temporal modeling and boundary probability modeling modules, respectively. From the results, we observe that both modules independently contribute to improving performance across all metrics. The combination of the two modules achieves the best results, demonstrating their complementary roles in mitigating boundary uncertainty and enhancing localization accuracy.
}

\begin{table}[t]
    \centering
    \caption{Results of different LLMs for query expansion with QD-DETR on the Charades-STA dataset.}
    \label{tab:llm}
    \begin{tabular}{l|ccc}
        \hline \hline
        \multicolumn{1}{c|}{LLM} & \multicolumn{1}{|c}{R1@0.5}  & \multicolumn{1}{c}{R1@0.7} & \multicolumn{1}{c}{mAP}\\ \midrule
        \multicolumn{1}{l|}{Noisy query} & 59.23 & 38.42 & 35.91\\
        \multicolumn{1}{l|}{Qwen} & 59.51 & 38.10 & 36.13\\
        \multicolumn{1}{l|}{LLaMa2-7B} & 59.46 & 38.28 & 35.81\\
        \multicolumn{1}{l|}{LLaMa2-13B} & 59.30 & 38.31 & \textbf{36.22}\\
        \multicolumn{1}{l|}{LLaMa3-8B} & \textbf{59.71} & \textbf{38.58} & 35.97\\
        \hline \hline
    \end{tabular}
\end{table}

\noindent \textbf{Evaluation of LLMs for Query Expansion.} 
To evaluate the impact of LLMs for query expansion on the performance, we use Qwen~\citep{bai2023qwen}, LLaMa2-7B~\citep{touvron2023llama}, LLaMa2-13B~\citep{touvron2023llama} for comparison, and the results are shown in
Table~\ref{tab:llm}. It is interesting to observe that different LLMs have little impact on performance, and all LLMs have similar R1@0.5, R1@0.7 and mAP scores. Among them, LLaMa3-8B achieves the best overall performance. 

In addition, we evaluate the robustness of our method to noisy expanded queries by randomly swapping the start and end queries for a portion of the samples. The results under the ``Noisy query'' setting show only a slight drop in performance, suggesting that our method remains stable even when the expanded queries contain a certain degree of noise.

\begin{table}[h]
\centering
\xiugai{
\caption{Ablation study of different prompt designs on TACoS dataset. ``Ours'' denotes the default prompt used in the paper.}
\label{tab:abl_prompt}
\begin{tabular}{l|ccc}
\hline \hline
Prompt Variant & R1@0.5 & R1@0.7 & mAP \\ \hline
Prompt-Object & 39.42 & 21.92 & 21.08 \\
Prompt-Concise & 40.61 & 23.99 & \textbf{22.76} \\
Ours & \textbf{40.83} & \textbf{24.04} & 22.49 \\
\hline \hline
\end{tabular}
}
\end{table}

\xiugai{
\noindent \textbf{Evaluation of Prompt for Query Expansion.} 
To evaluate the impact of prompt engineering on performance, we compare our default prompt with two variants: \textit{Prompt-Object}, which allows the LLM to freely describe objects without constraint, and \textit{Prompt-Concise}, which imposes a stricter length constraint of 10 words per sentence. The results are shown in Table~\ref{tab:abl_prompt}. It is observed that the inclusion of additional objects in \textit{Prompt-Object} leads to a performance drop, specifically decreasing R1@0.5 by 1.41\% and R1@0.7 by 2.12\%. This suggests that hallucinated objects not present in the original action introduce semantic noise that harms cross-modal alignment. Furthermore, the results of \textit{Prompt-Concise} show that further restricting the verbosity yields comparable results, indicating that our default 20-word constraint already provides an optimal balance between descriptive detail and noise control. These findings demonstrate that our prompt design effectively minimizes textual noise while preserving essential temporal cues.
}


\begin{figure}[t]
\centering  
    \includegraphics[width=0.9\linewidth]{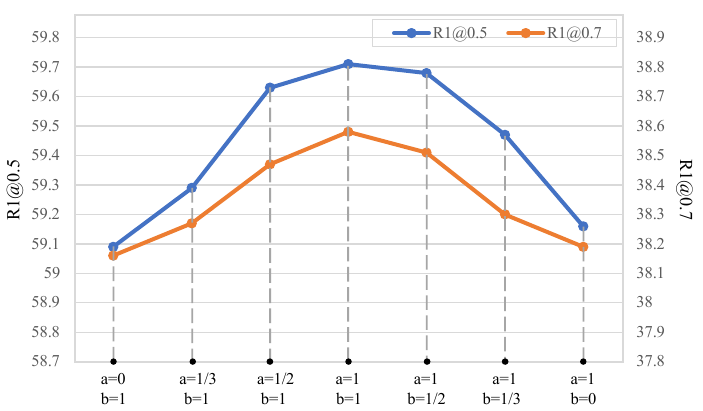} 
    \caption{Results of different values of hyperparameters ($a$ and $b$) for the global and local branches with QD-DETR on the Charades-STA dataset.}
    \label{fig:para}
\end{figure}

\begin{figure}[h]
    \centering
    \xiugai{
    \includegraphics[width=0.9\linewidth]{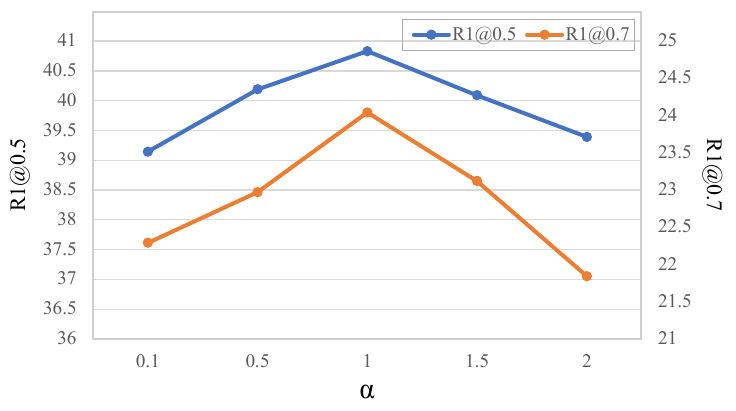}
    \caption{Results of different values of the hyperparameter ($\alpha$) for the global and local branches with QD-DETR on the Charades-STA dataset.}
    \label{fig:alpha}
    }
\end{figure}


\noindent \textbf{Evaluation of Hyperparameters in Temporal Modeling.} 
To analyze the effectiveness of the global and local branches in the query-guided temporal modeling module, conduct experiments using different weight parameter settings in Eq.(\ref{equ:enhance}), i.e., different values of hyperparameters $a$ and $b$. The results are shown in Figure~\ref{fig:para}. We observe that the combination of both global and local branches with equal weights ($a=b=1$) yields the best performance. When the weights are unbalanced, the performance drops slightly. In addition, when only one branch is used, the performance tends to degrade.

\begin{table*}[t]
    \centering
    \xiugai{
    \caption{Results of different boundary probability modeling strategies with QD-DETR on the Charades-STA and TACoS datasets.}
    \label{tab:prob}
    \begin{tabular}{l|ccc|ccc}
        \hline \hline
        \multicolumn{1}{c|}{} & \multicolumn{3}{c|}{Charades-STA} & \multicolumn{3}{c}{TACoS} \\
        \multicolumn{1}{c|}{Probability Modeling} & R1@0.5 & R1@0.7 & mAP & R1@0.5 & R1@0.7 & mAP\\ \hline
        Gauss distribution & 59.19 & 37.98 & 35.04 & 39.07 & 22.84 & 21.69 \\
        Only distance      & 59.11 & 37.34 & 35.44 & 40.19 & 23.09 & 21.70 \\
        Only similarity    & 59.35 & 38.04 & 35.33 & 39.64 & 22.94 & 21.43 \\
        Original query     & 59.35 & 38.33 & 35.61 & 39.67 & 23.22 & 21.87 \\
        \textbf{Ours}      & \textbf{59.71} & \textbf{38.58} & \textbf{35.97} & \textbf{40.83} & \textbf{24.04} & \textbf{22.49} \\
        \hline \hline
    \end{tabular}
    }
\end{table*}

\noindent \textbf{Evaluation of Boundary Probability Modeling.} To evaluate the effectiveness of the proposed boundary probability modeling strategy, we design four variants for comparison: (1) ``Gauss distribution", which directly uses Gauss distribution to smooth the boundary annotation; (2) ``Only distance", which only considers the temporal distance when calculating frame scores; (3) ``Only similarity", which only considers the semantic similarity when calculating frame scores; (4) ``Original query", which uses the original query to calculate similarity with video features when calculating frame scores.
\xiugai{Table~\ref{tab:prob} shows the proposed strategy achieves the best results across multiple datasets, which suggests the effectiveness of introducing both the temporal distance and semantic similarity to the boundary probability modeling. Notably, since the query expansion is performed as an offline preprocessing step and cached for training, our module provides performance gains with negligible impact on training efficiency and additional inference latency.}

\noindent \xiugai{\textbf{Evaluation of Hyperparameters in Boundary Probability Modeling.} We investigate the sensitivity of the model to the balancing coefficient $\alpha$ in Eq.(\ref{equ:pseudo}) and Eq.(\ref{equ:score_start}), which controls the relative importance of temporal distance versus semantic similarity during pseudo boundary generation. As shown in Figure~\ref{fig:alpha}, a smaller $\alpha$ (e.g., 0.5) relies more heavily on semantic alignment, which may lead to the selection of frames that are semantically similar but temporally distant from the ground truth, potentially introducing outliers. Conversely, a larger $\alpha$ (e.g., 1.5) imposes a stricter temporal constraint, which might limit the model's ability to adaptively find the true action boundaries when human annotations are imprecise. Experimental results indicate that $\alpha = 1.0$ provides the most robust performance, effectively balancing the global temporal prior with local semantic cues.}

\begin{figure}[t]
\centering  
    \includegraphics[width=0.9\linewidth]{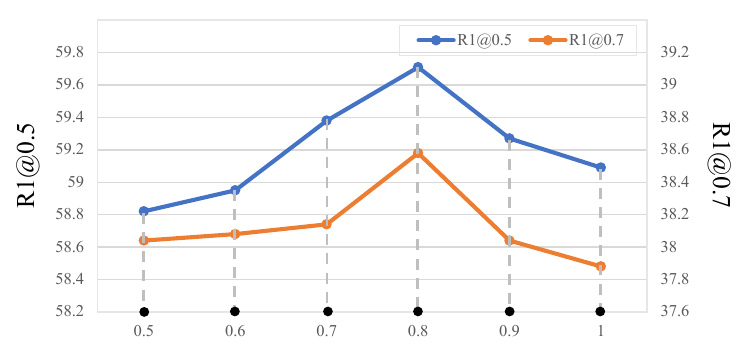} 
    \caption{Results of different values of the hyperparamete $\tau$.}
    \label{fig:tau}
\end{figure}

We also conduct experiments to analyze the impact of different $\tau$ values on model performance, as shown in Figure~\ref{fig:tau}. A lower $\tau$ (e.g., 0.5) results in more frames being included, which may introduce more noise. In contrast, a higher $\tau$ (e.g., 1.0)  filters out low-confidence frames, reducing noise but potentially missing relevant boundary frames. The results suggest that an intermediate value $\tau = 0.8$ achieves the optimal balance, offering the best performance.

\begin{figure}[t]
\centering  
\xiugai{
    \includegraphics[width=0.9\linewidth]{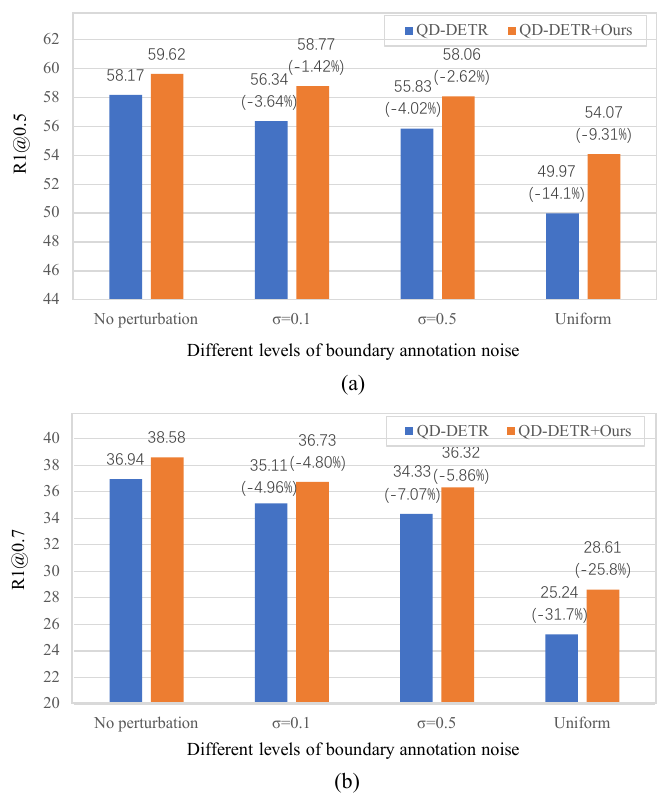} 
    \caption{Results of boundary annotation perturbation on the Charades-STA dataset by integrating our modules into QD-DETR.}
    \label{fig:label}
}
\end{figure}

\noindent \textbf{Evaluation of Robustness to Boundary Uncertainty.} 
To evaluate the robustness of our method to boundary uncertainty, we introduce perturbations to the boundary annotations in the Charades-STA training set to increase the boundary uncertainty during training. The original boundary annotations $\tau_s$, $\tau_e$ are perturbed to generate new annotations $\hat{\tau}_s$, $\hat{\tau}_e$, denoted by $\hat{\tau}_s=\tau_s + (\tau_e-\tau_s)\cdot X$, $\hat{\tau}_e=\tau_e + (\tau_e-\tau_s)\cdot Y$, where $X$ and $Y$ represent the distributions. 
We use three distributions for comparison: (1) ``$\sigma=0.1$", where $X, Y\sim N(0,0.1^2)$; (2) ``$\sigma=0.5$", where $X, Y\sim N(0,0.5^2)$; (3) ``Uniform", where $X, Y\sim U(-0.5,0.5)$. Note that these three distributions represent gradually increasing boundary annotation noise, which means that the uncertainty of boundary annotation increases.
\xiugai{
Figure~\ref{fig:label} shows that compared to QD-DETR, our method exhibits a relatively smaller performance drop as the level of annotation noise increases. Specifically, under a moderate perturbation of $\sigma=0.5$, the R1@0.5 of base QD-DETR drops by 4.02\%, while our framework only shows a 2.62\% decrease. This gap becomes even more pronounced under the most severe \textit{Uniform} perturbation, where the base model experiences a significant 14.10\% drop in R1@0.5, whereas our method restricts this decay to 9.31\%. A similar trend is observed in R1@0.7, where our model reduces the maximum drop from 31.67\% to 25.84\%.  This demonstrates the robustness of our modules to boundary uncertainty, as they maintain a more stable performance profile even when the grounding supervision is highly noisy.
}



\begin{table*}[h]
\centering
\xiugai{
\caption{\centering Performance comparison on different duration subsets on TACoS.}
\label{tab:subset_analysis_tacos}
\begin{tabular}{c|c|c|c|c|c}
\hline \hline
Subset & Size & Method & R1@0.5 & R1@0.7 & mAP \\
\hline
\multirow{2}{*}{Short ($< 6.8$s)} 
& \multirow{2}{*}{998} 
& QD-DETR   & 37.68 & 18.64 & 17.37 \\
& 
& +Ours  & 39.88 & 19.54 & 18.46 \\
\hline
\multirow{2}{*}{Average} 
& \multirow{2}{*}{2004} 
& QD-DETR   & 44.61 & 25.10 & 23.31 \\
& 
& +Ours  & 46.86 & 28.99 & 26.37 \\
\hline
\multirow{2}{*}{Long ($> 33.9$s)} 
& \multirow{2}{*}{999} 
& QD-DETR   & 26.13 & 12.41 & 16.62 \\
& 
& +Ours  & 30.13 & 14.51 & 18.30 \\
\hline
\hline
\end{tabular}
}
\end{table*}

\begin{table*}[h]
\centering
\xiugai{
\caption{\centering Performance comparison on different action location subsets on Charades-STA.}
\label{tab:subset_analysis_charades}
\begin{tabular}{c|c|c|c|c|c}
\hline \hline
Subset & Size & Method & R1@0.5 & R1@0.7 & mAP \\
\hline
\multirow{2}{*}{Beginning} 
& \multirow{2}{*}{957} 
& QD-DETR   & 70.85 & 48.90 & 44.42 \\
& 
& +Ours  & 71.16 & 51.93 & 45.25 \\
\hline
\multirow{2}{*}{End} 
& \multirow{2}{*}{665} 
& QD-DETR   & 62.41 & 46.47 & 42.09 \\
& 
& +Ours  & 63.46 & 47.37 & 42.49 \\
\hline
\hline
\end{tabular}
}
\end{table*}

\noindent \xiugai{
\textbf{Evaluation of Robustness to Action Duration and Location.} To further investigate the robustness of our framework across varying action durations and temporal positions, we conduct detailed subgroup analyses.
For the action duration analysis, as shown in Table~\ref{tab:subset_analysis_tacos}, we utilize the TACoS dataset due to its wide temporal span, categorizing samples in the test set into \textit{Short} (the shortest 25\%, $<6.8$s), \textit{Average}, and \textit{Long} (the longest 25\%, $>33.9$s) subsets. Our framework achieves consistent performance gains across all duration scales.
For the boundary-adjacent analysis, we evaluate our model on Charades-STA to leverage its high density of extremity-aligned samples. We define \textit{Beginning} samples as actions starting at 0s and \textit{End} samples as those ending within 1s of the video conclusion. As shown in Table~\ref{tab:subset_analysis_charades}, our method yields consistent improvements in both the \textit{Beginning} (+0.31\% R1@0.5) and \textit{End} (+1.05\% R1@0.5) scenarios. This confirms its ability to refine localization even when actions align with video limits.
}

\subsection{User Study on Quality of Generated Queries}

\begin{figure}[t]
\centering  
    \includegraphics[width=\linewidth]{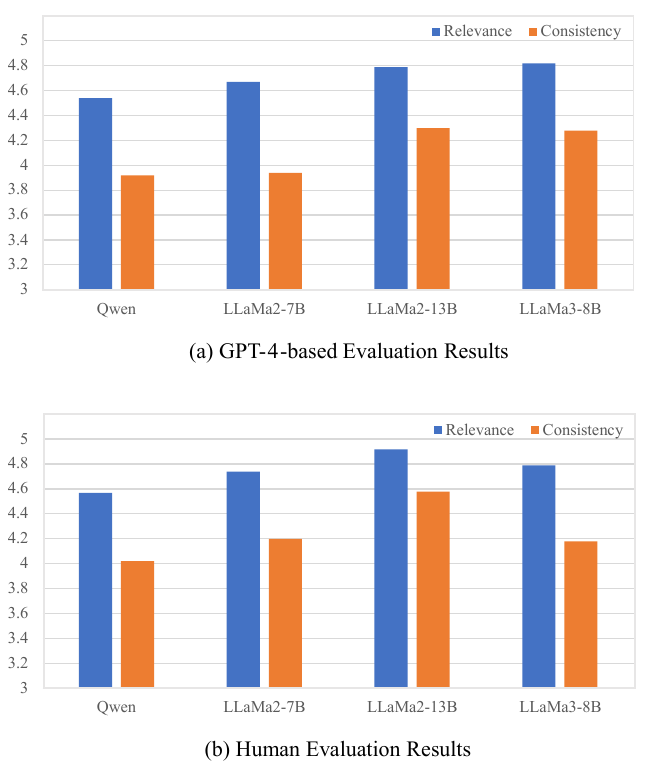} 
    \caption{Quantitative evaluation of query expansion quality by GPT-4 and human raters.}
    \label{fig:user_study}
\end{figure}

\xiugai{
To evaluate the quality and consistency of action queries generated by different LLMs, we conduct a user study using 100 sets of queries generated by each model (Qwen, LLaMa2-7B/13B, and LLaMa3-8B) on the Charades-STA dataset. Each set consists of the original query and its corresponding expanded boundary queries (start and end). 
We assess each set using both human and automated evaluations along two dimensions: (1) relevance to the original query in the dataset, and (2) consistency between the two generated queries. To ensure fairness and minimize bias, we anonymize the source model of each query set and randomize their presentation order. For the human evaluation, five participants with substantial experience in video-language research (but not involved in our project) rate each presented query set alongside the corresponding video frames, and the final score is calculated as the average of their ratings. For the automated evaluation, we prompt GPT-4 to score each query set on a scale of 0 (very poor) to 5 (excellent) for both relevance and consistency. 
}

As shown in Figure~\ref{fig:user_study}(a), GPT-4-based evaluation indicates that all models get relatively high scores, with larger models generally achieving higher relevance and consistency scores. The human evaluation results in Figure~\ref{fig:user_study}(b) reveal a similar pattern, indicating the agreement between automated and human assessments. These results demonstrate the overall high quality and consistency of the queries generated by our method.

\subsection{Qualitative Results}

\begin{figure*}[t]
\centering  
    \includegraphics[width=\linewidth]{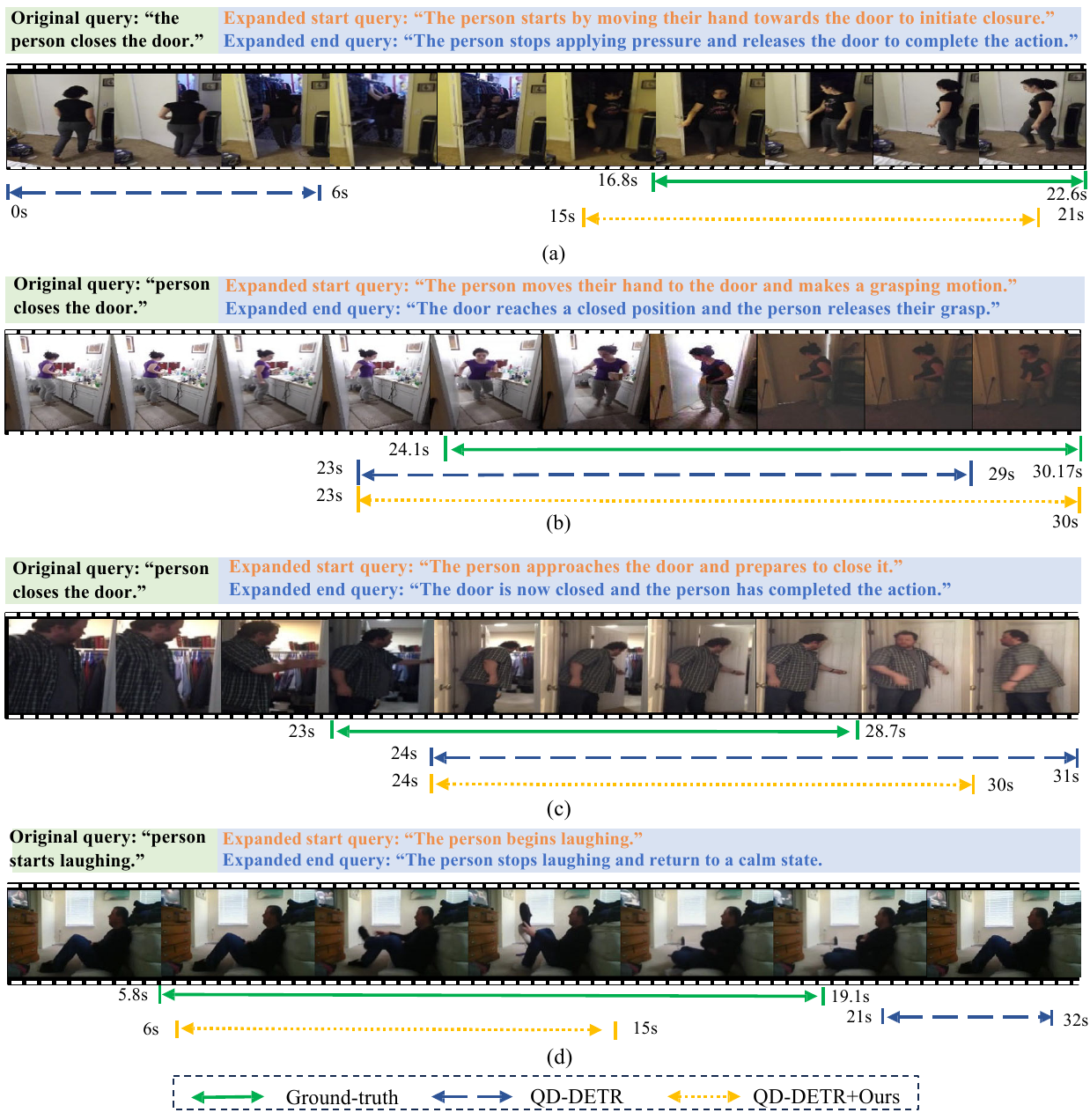} 
    \xiugai{
    \caption{Examples of action localization results with similar language queries on the Charades-STA dataset by integrating our modules into QD-DETR. The texts in red are the expanded start queries, and the texts in blue are the expanded end queries.}
    \label{fig:case}
    }
\end{figure*}
\subsubsection{Consistent Localization across Similar Queries}

To qualitatively evaluate the effectiveness of our modules, we presents several examples of action localization results by integrating our modules into QD-DETR on the Charades-STA dataset.
As shown in Figure~\ref{fig:case}, we observe that for similar queries of ``\textit{close the door}'', the start boundaries of the ground truth differ across examples: in Figure~\ref{fig:case} (a) and (b), the annotations are at ``\textit{starting to pull the door}'', while in Figure~\ref{fig:case} (c), the annotation is at ``\textit{reaching out to touch the door}''.
In contrast, we generate consistent expanded starting queries describing ``\textit{reaching out to touch the door}'' for three examples.
Consequently, our method predicts start boundaries consistently across examples, localizing them at ``\textit{reaching out to touch the door}'', which also outperforms the predictions of QD-DETR.

\subsubsection{Failure Case}

In Figure~\ref{fig:case}(d), with the query of “person starts laughing”, the action involves only subtle facial or body movements, making the transition into the action visually ambiguous. As a result, our model struggles to accurately determine the precise start point, highlighting the challenge posed by inherently ambiguous action boundaries.


\begin{figure*}[t]
\centering  
    \includegraphics[width=\linewidth]{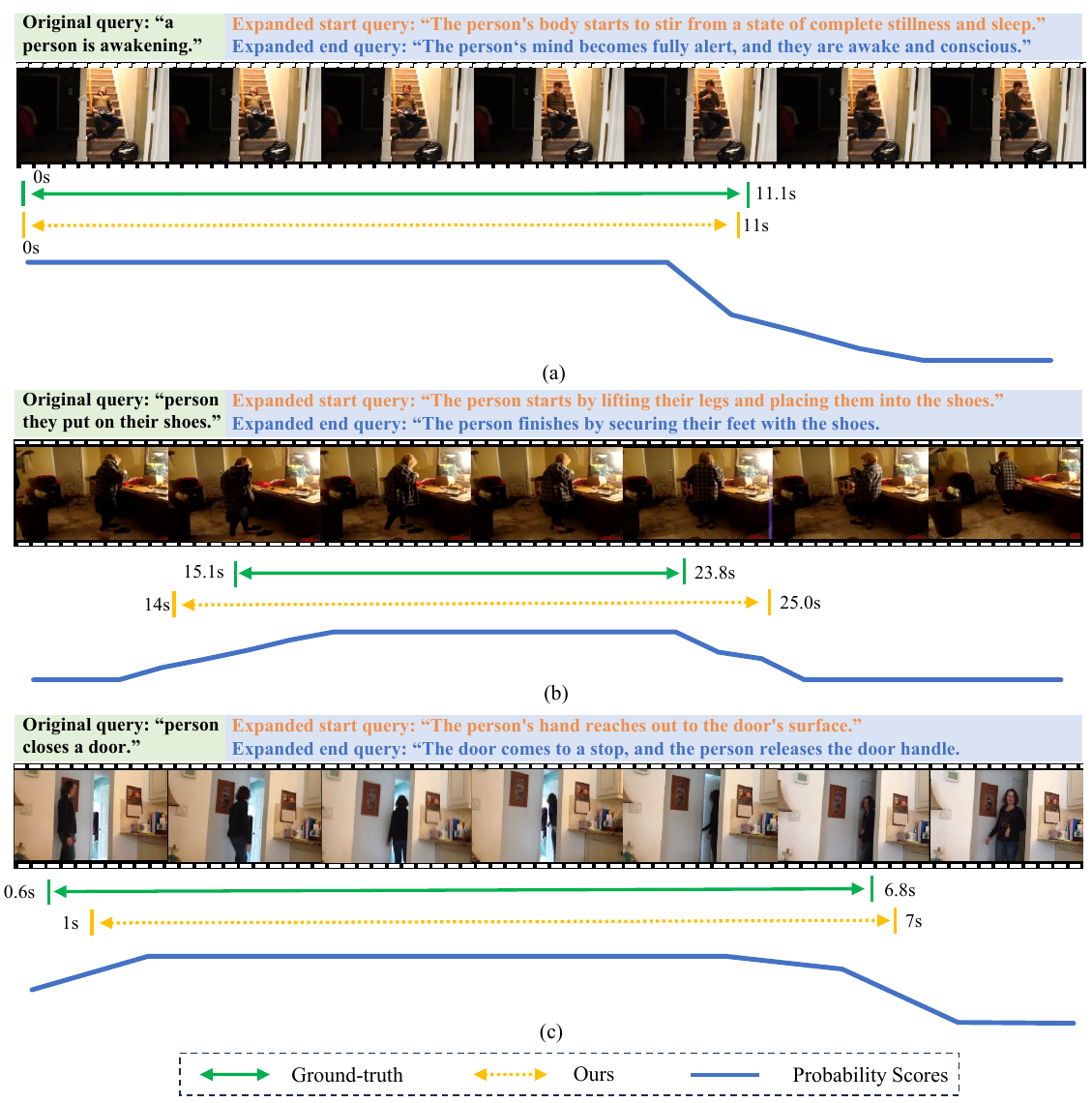} 
    \caption{Visualization of boundary probability scores on the Charades-STA dataset by integrating our modules into QD-DETR.}
    \label{fig:vis_prob}
\end{figure*}

\subsubsection{Visualization of Boundary Probability Scores}

To intuitively understand how our method models boundary uncertainty, we visualize the predicted boundary probability scores during training. As shown in Figure~\ref{fig:vis_prob}, our method produces meaningful probability peaks at frames corresponding to the \xiugai{action boundaries} described in the expanded queries, such as ``\textit{lifting their legs}''.
This indicates that the modeled probability scores effectively highlight the start and end of the action as guided by the expanded queries, providing robust boundary annotations.

\begin{table}[h]
\centering
\xiugai{
\caption{Inference latency and computational cost (per sample) on the TACoS dataset.}
\label{tab:latency_cost}
\begin{tabular}{c|c|c}
\hline \hline
Process & Time (ms) & FLOPs (G) \\
\hline
Query Expansion & $\sim$1315 & 1442.7 \\
\hline
Visual Encoder & $\sim$1344 & 54.9 \\
Text Encoder & $\sim$2.90 & 1.38 \\
QTM & $\sim$19.83 & 3.09 \\
Localization Head & $\sim$17.87 & 2.46 \\
\hline
\hline
\end{tabular}
}
\end{table}

\xiugai{
\subsection{Efficiency Analysis}}

\xiugai{To provide a comprehensive view for cold-start scenarios where real-time expansion is required, we report the end-to-end cost on TACoS in Table~\ref{tab:latency_cost}. The results show that while the LLM expansion (LLaMa3-8B) introduces an initial computation, its impact on the overall end-to-end latency is negligible. Crucially, the Query Expansion and Visual Encoding processes are executed in parallel. For a video with the average duration of the TACoS dataset ($\sim$4.7 minutes), the visual encoding (SlowFast+CLIP) takes approximately 1.34s on an RTX 4090 (averaging 9.53ms per 2s clip), which is comparable to the LLM's expansion time ($\sim$1.31s). Thus, the expansion process is effectively 'hidden' within the video processing pipeline, incurring zero additional time cost to the total system latency. In this context, the one-time semantic enrichment provided by the LLM is a worthwhile trade-off for the substantial localization gains.}

\section{Conclusion}\label{sec5}

We have presented a large language model (LLM)-powered query expansion method to enhance boundary prediction for language-driven action localization. 
By leveraging LLMs to generate expanded boundary queries that provide cues for start and end boundaries, our method captures boundary information that original language queries overlook. Additionally, the boundary probability modeling transforms rigid boundary annotations into probability scores, leveraging the semantic similarities between frames and the expanded query, as well as the temporal distances between frames and the annotated boundary frames, to provide more flexible and consistent supervision. 
The proposed modules are model-agnostic and can be seamlessly integrated into existing models of language-driven action localization in an off-the-shelf manner. 
Extensive experiments on five state-of-the-art models across three datasets validate the effectiveness of our method in reducing the impact of boundary uncertainty and enhancing boundary prediction.

\xiugai{
\section*{Declarations}
}

\noindent  \xiugai{\textbf{Ethical Approval} Not applicable.}

\noindent  \xiugai{\textbf{Conflict of Interest} The authors have no competing interests to declare that are relevant to the content of this article.}

\noindent  \xiugai{\textbf{Funding Declaration} This work was supported by the Shenzhen Science and Technology Program under Grant No. JCYJ20241202130548062, Guangdong Provincial Key Area Project of General Universities under Grant No.2025ZDZX3049 and Grant No.2024ZDZX1017.}

\noindent  \xiugai{\textbf{Data Availability} All datasets used in this study are open access and have been cited in the paper, i.e. TACoS (\url{https://www.coli.uni-saarland.de/projects/smile/page.php?id=tacos}), QVHighlights (\url{https://github.com/jayleicn/moment_detr}) and Charades-STA (\url{https://prior.allenai.org/projects/charades}).}

\noindent  \xiugai{\textbf{Consent for Publication } Not applicable.}

\noindent  \xiugai{\textbf{Author contribution } Shuo Yang is the corresponding author. The individual contributions are as follows:
\textit{Conceptualization:} Zirui Shang, Xinxiao Wu, and Shuo Yang. 
\textit{Methodology and Software:} Zirui Shang. 
\textit{Formal Analysis and Investigation:} Zirui Shang. 
\textit{Writing - Original Draft:} Zirui Shang. 
\textit{Writing - Review \& Editing:} Xinxiao Wu and Shuo Yang. 
\textit{Supervision:} Xinxiao Wu and Shuo Yang.
\textit{Funding Acquisition:} Xinxiao Wu and Shuo Yang.}

\noindent  \xiugai{\textbf{Consent to Participate } Informed consent was obtained from all individual participants involved in the user study. All participants were fully informed of the study's purpose and participated voluntarily.}

\backmatter

\bibliography{sn-bibliography}

@article{bai2023qwen,
 author = {Bai, Jinze and Bai, Shuai and Chu, Yunfei and Cui, Zeyu and Dang, Kai and Deng, Xiaodong and Fan, Yang and Ge, Wenbin and Han, Yu and Huang, Fei and others},
 journal = {ArXiv preprint},
 title = {Qwen technical report},
 year = {2023}
}

@article{carion2020end,
 author = {Carion, Nicolas and Massa, Francisco and Synnaeve, Gabriel and Usunier, Nicolas and Kirillov, Alexander and Zagoruyko, Sergey},
 journal = {Computer Vision--ECCV 2020: 16th European Conference on Computer Vision (ECCV)},
 pages = {213--229},
 title = {End-to-End Object Detection with Transformers},
 year = {2020}
}

@inproceedings{chen2019semantic,
 author = {Shaoxiang Chen and
Yu{-}Gang Jiang},
 booktitle = {Proc. of AAAI},
 pages = {8199--8206},
 title = {Semantic Proposal for Activity Localization in Videos via Sentence
Query},
 year = {2019}
}

@article{dubey2024llama,
 author = {Dubey, Abhimanyu and Jauhri, Abhinav and Pandey, Abhinav and Kadian, Abhishek and Al-Dahle, Ahmad and Letman, Aiesha and Mathur, Akhil and Schelten, Alan and Yang, Amy and Fan, Angela and others},
 journal = {ArXiv preprint},
 title = {The llama 3 herd of models},
 year = {2024}
}

@inproceedings{feichtenhofer2019slowfast,
 author = {Christoph Feichtenhofer and
Haoqi Fan and
Jitendra Malik and
Kaiming He},
 booktitle = {Proc. of ICCV},
 pages = {6201--6210},
 title = {SlowFast Networks for Video Recognition},
 year = {2019}
}

@inproceedings{gao2017tall,
 author = {Jiyang Gao and
Chen Sun and
Zhenheng Yang and
Ram Nevatia},
 booktitle = {Proc. of ICCV},
 pages = {5277--5285},
 title = {{TALL:} Temporal Activity Localization via Language Query},
 year = {2017}
}

@inproceedings{gao2021fast,
 author = {Junyu Gao and
Changsheng Xu},
 booktitle = {Proc. of ICCV},
 pages = {1503--1512},
 title = {Fast Video Moment Retrieval},
 year = {2021}
}

@inproceedings{gao2021relation,
 author = {Gao, Jialin  and
Sun, Xin  and
Xu, Mengmeng  and
Zhou, Xi  and
Ghanem, Bernard},
 booktitle = {Proc. of EMNLP},
 pages = {3978--3988},
 title = {Relation-aware Video Reading Comprehension for Temporal Language Grounding},
 year = {2021}
}

@inproceedings{ghosh2019excl,
 author = {Ghosh, Soham  and
Agarwal, Anuva  and
Parekh, Zarana  and
Hauptmann, Alexander},
 booktitle = {Proc. of NAACL},
 pages = {1984--1990},
 title = {{E}x{CL}: {E}xtractive {C}lip {L}ocalization {U}sing {N}atural {L}anguage {D}escriptions},
 year = {2019}
}

@inproceedings{hendricks2017localizing,
 author = {Lisa Anne Hendricks and
Oliver Wang and
Eli Shechtman and
Josef Sivic and
Trevor Darrell and
Bryan C. Russell},
 booktitle = {Proc. of ICCV},
 pages = {5804--5813},
 title = {Localizing Moments in Video with Natural Language},
 year = {2017}
}

@inproceedings{huang2022video,
 author = {Huang, Jiabo and Jin, Hailin and Gong, Shaogang and Liu, Yang},
 booktitle = {Proc. of ECCV},
 pages = {724--740},
 title = {Video activity localisation with uncertainties in temporal boundary},
 year = {2022}
}

@inproceedings{jang2023knowing,
 author = {Jinhyun Jang and
Jungin Park and
Jin Kim and
Hyeongjun Kwon and
Kwanghoon Sohn},
 booktitle = {Proc. of ICCV},
 pages = {13800--13810},
 title = {Knowing Where to Focus: Event-aware Transformer for Video Grounding},
 year = {2023}
}

@inproceedings{lei2021detecting,
 author = {Jie Lei and
Tamara L. Berg and
Mohit Bansal},
 booktitle = {Proc. of NeurIPS},
 pages = {11846--11858},
 title = {Detecting Moments and Highlights in Videos via Natural Language Queries},
 year = {2021}
}

@inproceedings{liu2021adaptive,
 author = {Liu, Daizong  and
Qu, Xiaoye  and
Dong, Jianfeng  and
Zhou, Pan},
 booktitle = {Proc. of EMNLP},
 pages = {9292--9301},
 title = {Adaptive Proposal Generation Network for Temporal Sentence Localization in Videos},
 year = {2021}
}

@inproceedings{liu2021context,
 author = {Daizong Liu and
Xiaoye Qu and
Jianfeng Dong and
Pan Zhou and
Yu Cheng and
Wei Wei and
Zichuan Xu and
Yulai Xie},
 booktitle = {Proc. of CVPR},
 pages = {11235--11244},
 title = {Context-Aware Biaffine Localizing Network for Temporal Sentence Grounding},
 year = {2021}
}

@inproceedings{liu2021progressively,
 author = {Liu, Daizong  and
Qu, Xiaoye  and
Zhou, Pan},
 booktitle = {Proc. of EMNLP},
 pages = {9302--9311},
 title = {Progressively Guide to Attend: An Iterative Alignment Framework for Temporal Sentence Grounding},
 year = {2021}
}

@inproceedings{liu2022memory,
 author = {Daizong Liu and
Xiaoye Qu and
Xing Di and
Yu Cheng and
Zichuan Xu and
Pan Zhou},
 booktitle = {Proc. of AAAI},
 pages = {1665--1673},
 title = {Memory-Guided Semantic Learning Network for Temporal Sentence Grounding},
 year = {2022}
}

@inproceedings{liu2022reducing,
 author = {Liu, Daizong and Qu, Xiaoye and Hu, Wei},
 booktitle = {Proc. of ACM MM},
 pages = {4092--4101},
 title = {Reducing the vision and language bias for temporal sentence grounding},
 year = {2022}
}

@inproceedings{liu2022skimming,
 author = {Liu, Daizong and Hu, Wei},
 booktitle = {Proc. of ACM MM},
 pages = {4536--4545},
 title = {Skimming, Locating, then Perusing: A Human-like Framework for Natural Language Video Localization},
 year = {2022}
}

@inproceedings{liu2023jointly,
 author = {Daizong Liu and
Pan Zhou},
 booktitle = {Proc. of ICASSP},
 pages = {1--5},
 title = {Jointly Visual- and Semantic-Aware Graph Memory Networks for Temporal
Sentence Localization in Videos},
 year = {2023}
}

@inproceedings{liu2023towards,
 author = {Zhihang Liu and
Jun Li and
Hongtao Xie and
Pandeng Li and
Jiannan Ge and
Sun'ao Liu and
Guoqing Jin},
 booktitle = {Proc. of AAAI},
 pages = {3855--3863},
 title = {Towards Balanced Alignment: Modal-Enhanced Semantic Modeling for Video
Moment Retrieval},
 year = {2024}
}

@article{moon2023correlation,
 author = {Moon, WonJun and Hyun, Sangeek and Lee, SuBeen and Heo, Jae-Pil},
 journal = {ArXiv preprint},
 title = {Correlation-guided query-dependency calibration in video representation learning for temporal grounding},
 year = {2023}
}

@inproceedings{moon2023query,
 author = {WonJun Moon and
Sangeek Hyun and
Sanguk Park and
Dongchan Park and
Jae{-}Pil Heo},
 booktitle = {Proc. of CVPR},
 pages = {23023--23033},
 title = {Query - Dependent Video Representation for Moment Retrieval and Highlight
Detection},
 year = {2023}
}

@inproceedings{mun2020local,
 author = {Jonghwan Mun and
Minsu Cho and
Bohyung Han},
 booktitle = {Proc. of CVPR},
 pages = {10807--10816},
 title = {Local-Global Video-Text Interactions for Temporal Grounding},
 year = {2020}
}

@inproceedings{otani2020uncovering,
 author = {Mayu Otani and
Yuta Nakashima and
Esa Rahtu and
Janne Heikkil{\"{a}}},
 booktitle = {Proc. of BMVC},
 title = {Uncovering Hidden Challenges in Query-Based Video Moment Retrieval},
 year = {2020}
}

@article{pan2022video,
 author = {Pan, Wenwen and Zhao, Zhou and Huang, Wencan and Zhang, Zhu and Fu, Liyong and Pan, Zhigeng and Yu, Jun and Wu, Fei},
 journal = {IEEE Transactions on Neural Networks and Learning Systems},
 title = {Video moment retrieval with noisy labels},
 year = {2022}
}

@inproceedings{radford2021learning,
 author = {Alec Radford and
Jong Wook Kim and
Chris Hallacy and
Aditya Ramesh and
Gabriel Goh and
Sandhini Agarwal and
Girish Sastry and
Amanda Askell and
Pamela Mishkin and
Jack Clark and
Gretchen Krueger and
Ilya Sutskever},
 booktitle = {Proc. of ICML},
 pages = {8748--8763},
 title = {Learning Transferable Visual Models From Natural Language Supervision},
 year = {2021}
}

@article{regneri2013grounding,
 author = {Regneri, Michaela  and
Rohrbach, Marcus  and
Wetzel, Dominikus  and
Thater, Stefan  and
Schiele, Bernt  and
Pinkal, Manfred},
 journal = {TACL},
 pages = {25--36},
 title = {Grounding Action Descriptions in Videos},
 year = {2013}
}

@inproceedings{rohrbach2012script,
 author = {Rohrbach, Marcus and Regneri, Michaela and Andriluka, Mykhaylo and Amin, Sikandar and Pinkal, Manfred and Schiele, Bernt},
 booktitle = {Proc. of ECCV},
 pages = {144--157},
 title = {Script data for Attribute-based Recognition of Composite Activities},
 year = {2012}
}

@inproceedings{sigurdsson2016hollywood,
 author = {Sigurdsson, Gunnar A and Varol, G{\"u}l and Wang, Xiaolong and Farhadi, Ali and Laptev, Ivan and Gupta, Abhinav},
 booktitle = {Proc. of ECCV},
 pages = {510--526},
 title = {Hollywood in Homes: Crowdsourcing Data Collection for Activity Understanding},
 year = {2016}
}

@article{sun2021video,
 author = {Sun, Guanglu and Liang, Lili and Li, Tianlin and Yu, Bo and Wu, Meng and Zhang, Bolun},
 journal = {Mobile Networks and Applications},
 pages = {1--34},
 title = {Video question answering: a survey of models and datasets},
 year = {2021}
}

@inproceedings{Sun2022YouNT,
 author = {Xin Sun and
Xuan Wang and
Jialin Gao and
Qiong Liu and
Xi Zhou},
 booktitle = {Proc. of SIGIR},
 pages = {1022--1032},
 title = {You Need to Read Again: Multi-granularity Perception Network for Moment
Retrieval in Videos},
 year = {2022}
}

@inproceedings{taichi2024emnlp,
 author = {Taichi Nishimura and Shota Nakada and Hokuto Munakata and Tatsuya Komatsu},
 booktitle = {Proc. of EMNLP},
 title = {Lighthouse: A User-Friendly Library for Reproducible Video Moment Retrieval and Highlight Detection},
 year = {2024}
}

@article{tang2021frame,
 author = {Tang, Haoyu and Zhu, Jihua and Liu, Meng and Gao, Zan and Cheng, Zhiyong},
 journal = {IEEE Transactions on Multimedia},
 pages = {1338--1349},
 title = {Frame-wise cross-modal matching for video moment retrieval},
 year = {2021}
}

@article{touvron2023llama,
 author = {Touvron, Hugo and Lavril, Thibaut and Izacard, Gautier and Martinet, Xavier and Lachaux, Marie-Anne and Lacroix, Timoth{\'e}e and Rozi{\`e}re, Baptiste and Goyal, Naman and Hambro, Eric and Azhar, Faisal and others},
 journal = {ArXiv preprint},
 title = {Llama: Open and Efficient Foundation Language Models},
 year = {2023}
}

@article{wang2022cross,
 author = {Wang, Gongmian and Xu, Xing and Shen, Fumin and Lu, Huimin and Ji, Yanli and Shen, Heng Tao},
 journal = {IEEE Transactions on Multimedia},
 pages = {1221--1232},
 title = {Cross-modal Dynamic Networks for Video Moment Retrieval with Text Query},
 year = {2022}
}

@inproceedings{wu2021diving,
 author = {Wu, Ziyue and Gao, Junyu and Huang, Shucheng and Xu, Changsheng},
 booktitle = {Proc. of ICME},
 pages = {1--6},
 title = {Diving into the Relations: Leveraging Semantic and Visual Structures for Video Moment Retrieval},
 year = {2021}
}

@inproceedings{xiao2024bridging,
 author = {Xiao, Yicheng and Luo, Zhuoyan and Liu, Yong and Ma, Yue and Bian, Hengwei and Ji, Yatai and Yang, Yujiu and Li, Xiu},
 booktitle = {Proc. of CVPR},
 pages = {18709--18719},
 title = {Bridging the gap: A unified video comprehension framework for moment retrieval and highlight detection},
 year = {2024}
}

@inproceedings{xu2017r,
 author = {Huijuan Xu and
Abir Das and
Kate Saenko},
 booktitle = {Proc. of ICCV},
 pages = {5794--5803},
 title = {{R-C3D:} Region Convolutional 3D Network for Temporal Activity Detection},
 year = {2017}
}

@inproceedings{xu2019multilevel,
 author = {Huijuan Xu and
Kun He and
Bryan A. Plummer and
Leonid Sigal and
Stan Sclaroff and
Kate Saenko},
 booktitle = {Proc. of AAAI},
 pages = {9062--9069},
 title = {Multilevel Language and Vision Integration for Text-to-Clip Retrieval},
 year = {2019}
}

@inproceedings{yang2024task,
 author = {Yang, Jin and Wei, Ping and Li, Huan and Ren, Ziyang},
 booktitle = {Proc. of CVPR},
 pages = {18308--18318},
 title = {Task-Driven Exploration: Decoupling and Inter-Task Feedback for Joint Moment Retrieval and Highlight Detection},
 year = {2024}
}

@inproceedings{yuan2019semantic,
 author = {Yitian Yuan and
Lin Ma and
Jingwen Wang and
Wei Liu and
Wenwu Zhu},
 booktitle = {Proc. of NeurIPS},
 pages = {534--544},
 title = {Semantic Conditioned Dynamic Modulation for Temporal Sentence Grounding
in Videos},
 year = {2019}
}

@inproceedings{zeng2021multi,
 author = {Yawen Zeng and
Da Cao and
Xiaochi Wei and
Meng Liu and
Zhou Zhao and
Zheng Qin},
 booktitle = {Proc. of CVPR},
 pages = {2215--2224},
 title = {Multi-Modal Relational Graph for Cross-Modal Video Moment Retrieval},
 year = {2021}
}

@inproceedings{zhang2020learning,
 author = {Songyang Zhang and
Houwen Peng and
Jianlong Fu and
Jiebo Luo},
 booktitle = {Proc. of AAAI},
 pages = {12870--12877},
 title = {Learning 2D Temporal Adjacent Networks for Moment Localization with
Natural Language},
 year = {2020}
}

@inproceedings{zhang2020span,
 author = {Zhang, Hao  and
Sun, Aixin  and
Jing, Wei  and
Zhou, Joey Tianyi},
 booktitle = {Proc. of ACL},
 pages = {6543--6554},
 title = {Span-based Localizing Network for Natural Language Video Localization},
 year = {2020}
}

@article{zhang2021natural,
 author = {Zhang, Hao and Sun, Aixin and Jing, Wei and Zhen, Liangli and Zhou, Joey Tianyi and Goh, Rick Siow Mong},
 journal = {IEEE Transactions on Pattern Analysis and Machine Intelligence},
 pages = {4252-4266},
 title = {Natural Language Video Localization: A Revisit in Span-Based Question Answering Framework},
 year = {2022}
}

@article{zhang2022ai,
 author = {Zhang, Xinrong and Li, Yanghao and Han, Yuxing and Wen, Jiangtao},
 title = {AI video editing: A survey},
 journal = {Preprints.org},
 year = {2022}
}

@inproceedings{zhang2023text,
 author = {Yimeng Zhang and
Xin Chen and
Jinghan Jia and
Sijia Liu and
Ke Ding},
 booktitle = {Proc. of CVPR},
 pages = {14794--14804},
 title = {Text-Visual Prompting for Efficient 2D Temporal Video Grounding},
 year = {2023}
}

@inproceedings{jung2025background,
  title={Background-aware moment detection for video moment retrieval},
  author={Jung, Minjoon and Jang, Youwon and Choi, Seongho and Kim, Joochan and Kim, Jin-Hwa and Zhang, Byoung-Tak},
  booktitle={2025 IEEE/CVF Winter Conference on Applications of Computer Vision (WACV)},
  pages={8586--8596},
  year={2025},
  organization={IEEE}
}

@article{flanagan2025moment,
  title={Moment of Untruth: Dealing with Negative Queries in Video Moment Retrieval},
  author={Flanagan, Kevin and Damen, Dima and Wray, Michael},
  journal={arXiv preprint arXiv:2502.08544},
  year={2025}
}

@article{fang2025multi,
  title={Multi-modal integrated proposal generation network for weakly supervised video moment retrieval},
  author={Fang, Dikai and Xu, Huahu and Wei, Wei and Guizani, Mohsen and Gao, Honghao},
  journal={Expert Systems with Applications},
  volume={269},
  pages={126497},
  year={2025},
  publisher={Elsevier}
}

@article{huo2025skim,
  title={Skim-and-scan transformer: A new transformer-inspired architecture for video-query based video moment retrieval},
  author={Huo, Shuwei and Zhou, Yuan and Chen, Keran and Xiang, Wei},
  journal={Expert Systems with Applications},
  volume={270},
  pages={126525},
  year={2025},
  publisher={Elsevier}
}

@article{lv2025variational,
  title={Variational global clue inference for weakly supervised video moment retrieval},
  author={Lv, Zezhong and Su, Bing},
  journal={Knowledge-Based Systems},
  volume={311},
  pages={113071},
  year={2025},
  publisher={Elsevier}
}

@article{liu2025and,
  title={What and Where: Semantic Grasping and Contextual Scanning for Moment Retrieval and Highlight Detection},
  author={Liu, Jing and He, Zhuo and Nie, Weizhi and Zhang, Zongbing and Su, Yuting},
  journal={IEEE Transactions on Circuits and Systems for Video Technology},
  year={2025},
  publisher={IEEE}
}

\end{document}